%% file: main.tex
\documentclass[conference]{IEEEtran}
\IEEEoverridecommandlockouts
\usepackage{cite}
\usepackage{balance}
\usepackage{amsmath,amssymb,amsfonts}
\usepackage{algorithmic}
\usepackage{graphicx}
\usepackage{textcomp}
\usepackage[caption=false]{subfig}
\usepackage[table,xcdraw]{xcolor}
\usepackage[utf8]{inputenc} 
\usepackage[T1]{fontenc} 
\def\BibTeX{{\rm B\kern-.05em{\sc i\kern-.025em b}\kern-.08em
    T\kern-.1667em\lower.7ex\hbox{E}\kern-.125emX}}

\newcommand{\dd}{\mathrm{\,d}} 

\usepackage{subfig}%
\usepackage{tikz}
\usepackage{pgfplots}
\tikzset{every picture/.append style={font=\scriptsize}}

\usepackage{siunitx}

\begin{document}

\title{Spatio-Temporal Multisensor Calibration Based on Gaussian Processes Moving Object Tracking \\
}

\author{Juraj Peršić, Luka Petrović, Ivan Marković, Ivan Petrović$^\ast$
\thanks{
This work has been supported by the European Regional Development Fund under the project \textit{System for increased driving safety in public urban rail traffic} (SafeTRAM). This research has also been carried out within the activities of the \textit{Centre of Research Excellence for Data Science and Cooperative Systems} supported by the Ministry of Science and Education of the Republic of Croatia.
}
\thanks{
$^{\ast}$Authors are with the University of Zagreb
Faculty of Electrical Engineering and Computing, Laboratory for Autonomous Systems and Mobile Robotics, Croatia. {\{juraj.persic, luka.petrovic, ivan.markovic, ivan.petrovic\}@fer.hr }
}
}
\maketitle

\begin{abstract}

Perception is one of the key abilities of autonomous mobile robotic systems, which often relies on fusion of heterogeneous sensors.
Although this heterogeneity presents a challenge for sensor calibration, it is also the main prospect for reliability and robustness of autonomous systems.
In this paper, we propose a method for multisensor calibration based on Gaussian processes (GPs) estimated moving object trajectories, resulting with temporal and extrinsic parameters.
The appealing properties of the proposed temporal calibration method are: coordinate frame invariance, thus avoiding prior extrinsic calibration, theoretically grounded batch state estimation and interpolation using GPs, computational efficiency with $\mathcal{O}(n)$ complexity, leveraging data already available in autonomous robot platforms, and the end result enabling 3D point-to-point extrinsic multisensor calibration.
The proposed method is validated both in simulations and real-world experiments.
For real-world experiment we evaluated the method on two multisensor systems: an externally triggered stereo camera, thus having temporal ground truth readily available, and a heterogeneous combination of a camera and motion capture system.
The results show that the estimated time delays are accurate up to a fraction of the fastest sensor sampling time.
\end{abstract}


\section{Introduction}
\label{sec:intro}
\input{text/intro}

\vspace{0.3cm}
\section{Theoretical Background}
\label{sec:gaussian}
\input{text/gaussian.tex}
\vspace{0.3cm}
\section{Proposed Calibration Method}
\label{sec:method}
\input{text/method.tex}
\vspace{0.3cm}
\section{Experimental Results}
\label{sec:results}
\input{text/results.tex}

\vspace{0.3cm}
\section{Conclusion}
\label{sec:conclusion}
\input{text/conclusion.tex}

\vspace{0.3cm}
\bibliographystyle{IEEEtran}
\bibliography{library}

\end{document}

%% file: text/intro.tex
Modern autonomous robotic systems navigate through the environment using information gathered by various sensors.
To process the gathered information, robots must rely on accurate sensor models and often fuse information from multiple sensors to improve performance.
For sensor fusion, appropriate knowledge of both spatial and temporal relations between the sensors is required, which can be challenging when working with heterogenous sensor systems, since sensors can operate based on various physical phenomena, different frame rates, and even have non or barely overlapping field-of-views.
The previously described challenge is termed sensor calibration and can be divided into intrinsic, extrinsic, and temporal calibration.
The intrinsic calibration is related to individual sensors as it provides parameters for sensor models.
The task of the extrinsic calibration is to find homogeneous transforms relating multiple sensors, while temporal calibration aims to find time delays between the individual sensor clocks.

The sensor calibration approach for a particular problem depends on multiple factors, e.g., the type of involved sensors, overlapping field of view, required degree of calibration accuracy,
necessity for online recalibration etc.
Herein, we focus on the tracking-based and temporal calibration methods, and additionally emphasize approaches relying on continuous time representation using Gaussian processes (GPs).
The advantage of tracking-based approaches is that one can achieve temporal calibration by using just the coordinate frame invariant magnitudes of the tracked target velocities, thus avoiding the need for previous extrinsic calibration, and by leveraging information already available in multisensor platforms navigating in dynamic environments, e.g., autonomous vehicles and mobile robots.
Furthermore, leveraging GPs enables us a theoretically grounded batch state estimation and interpolation, being
a well recognized tool in machine learning \cite{Rasmussen2004} both for regression and classification problems, and have been proposed for a variety of robotics challenges as well \cite{Barfoot2017}.
%
%
For example, in \cite{Barfoot2014, Anderson2015} mobile robot localization was a motivation for an efficient batch state estimation using GP regression, in \cite{Mukadam2016} GPs have been used for efficient motion planning, being especially valuable in high-dimensional configuration spaces, while in \cite{Wahlstrom2015} they were used for tracking of extended targets.

 \begin{figure}[t]
\centering
    \def\svgwidth{1\columnwidth}
    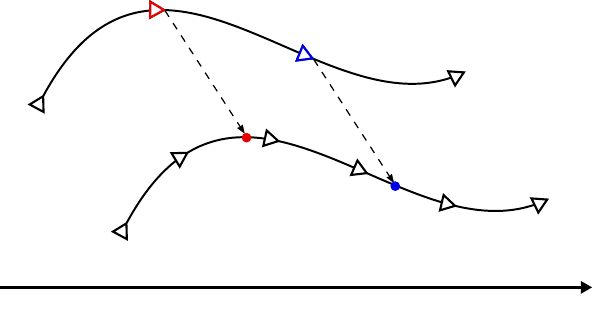
\caption{Continuous time trajectory representation using Gaussian processes provides an elegant temporal registration of asynchronous measurements. Illustration shows the time delay estimation by aligning two velocity magnitude trajectories, ${}^1v(t)$ and ${}^2v(t)$. States of the anchor sensor at measurement times (triangles) and states at interpolated times (circle) are used to generate correspondences (blue and red pairs). }
    \label{fig:temporal}
\end{figure}

To calibrate extrinsically and temporally multiple sensors, we need to perform correspondence registration in the sensor data, which is later used to create an optimization criterion.
The correspondences can originate from a designed target, yielding the target-based methods \cite{Richardson2013}, or from the environment itself, as in the case of the so called targetless methods \cite{Scott2016,Levinson2013}.
For example, odometry-based methods are a special class of targetless methods suitable for online application and are based on leveraging the environment to estimate ego-motion and calibrate the multisensor system \cite{Maye2015,Keivan2015}.
The concept of sensor calibration by aligning trajectories of moving objects received most attention in the target-based calibration of depth sensors \cite{Fornaser2017,Faion2015, Su2018}, and calibration of cameras, depth sensors, and LiDARs by exploiting human motion \cite{Lv2006,Glas2010,Glas2015, Tang2017}.
%
%
Specifically, to match trajectories between the sensors, the authors observe a similarity measure of the net velocity history profiles; however, in the optimization step, they rely only on the detected positions of the tracked people.
In \cite{Quenzel2016}, authors propose to calibrate multiple 2D LiDARs by tracking moving objects using a pose graph, wherein rotation is decoupled from translation by using a multiple rotation averaging approach.

Temporal calibration of a sensor system requires motion, either of the observed target \cite{Fornaser2017,Huber2014} or the system itself \cite{Kelly2014, Rehder2016,Li2014,Ackerman2013a}.
In \cite{Fornaser2017} authors tracked a colored sphere to perform spatio-temporal calibration of multiple Kinect v2 sensors.
By performing principal component analysis on the trajectories, they obtained field of view invariant one-dimensional kernels used in temporal calibration.
Even though this method is applicable to other sensors, it assumes the same frame rate of the sensors and its resolution is limited to the sampling time.
%
In \cite{Huber2014}, temporal calibration based on object tracking is presented where the author use linear interpolation for continuous time representation and position norm for the dimensionality reduction.
%
%
The \textit{$AX=XB$} sensor calibration problem was tackled \cite{Ackerman2013a} with unknown temporal correspondences.
To perform dimensionality reduction, authors used one-dimensional invariants -- displacement and angle of rotation -- defined by Pl{\"u}cker coordinates of the screw motion.
In \cite{Kelly2014}, authors proposed an algorithm based on system motion by aligning curves in the 3D orientation space.
The temporal calibration problem was formulated as a registration task which can be considered as a variant of the iterative closest point algorithm.
The approach in \cite{Rehder2016} targets the motion-based spatio-temporal calibration in the continuous-time domain using B-splines, having the availability of the analytical derivatives and effective state number reduction, but requiring setting the order and number of splines to  appropriately capture sensor dynamics.

In this paper, we propose a sensor calibration framework based on trajectories of the tracked target to generate correspondences between the sensors.
%
The advantages of the proposed temporal calibration method are (i) coordinate frame invariance, since we are using velocity magnitudes and thus avoiding the need for previous extrinsic calibration, (ii) theoretically grounded batch state estimation and interpolation, based on the theory of GPs, (iii) computational efficiency resulting with $\mathcal{O}(N)$ complexity with respect to the number of measurements, (iv) leveraging data already available in autonomous platforms, e.g., autonomous vehicles must track moving objects in the environment, and (v) the end result enabling extrinsic multisensor calibration using well known 3D point-to-point methods.
%
%
%
%
%
We evaluate the proposed method both in simulations and real-world experiments.
We conducted real-world experiments on two multisensor setups tracking a known calibration target: an externally triggered stereo camera, thus opportunely having zero delay as the available ground truth, and a heterogeneous setup consisting of a camera and a motion capture system.
Note that the proposed method \emph{requires only} that sensors can track the same moving object.
%
%
Furthermore, the paper is complemented by an open-source ROS toolbox \emph{Calirad} implementing the proposed method and a C++ library \emph{ESGPR} implementing the GP regression\footnote{bitbucket.org/unizg-fer-lamor}.
%

The rest of the paper is organized as follows.
Section~\ref{sec:gaussian} provides theoretical insights on the GP regression and the exactly sparse GP priors.
Section~\ref{sec:method} elaborates the proposed calibration method, while experiments are given in Section~\ref{sec:results}.
In the end, the paper is concluded in Section~\ref{sec:conclusion}.

%% file: figures/time_delay.pdf_tex
\begingroup%
  \makeatletter%
  \providecommand\color[2][]{%
    \errmessage{(Inkscape) Color is used for the text in Inkscape, but the package 'color.sty' is not loaded}%
    \renewcommand\color[2][]{}%
  }%
  \providecommand\transparent[1]{%
    \errmessage{(Inkscape) Transparency is used (non-zero) for the text in Inkscape, but the package 'transparent.sty' is not loaded}%
    \renewcommand\transparent[1]{}%
  }%
  \providecommand\rotatebox[2]{#2}%
  \newcommand*\fsize{\dimexpr\f@size pt\relax}%
  \newcommand*\lineheight[1]{\fontsize{\fsize}{#1\fsize}\selectfont}%
  \ifx\svgwidth\undefined%
    \setlength{\unitlength}{175.4452803bp}%
    \ifx\svgscale\undefined%
      \relax%
    \else%
      \setlength{\unitlength}{\unitlength * \real{\svgscale}}%
    \fi%
  \else%
    \setlength{\unitlength}{\svgwidth}%
  \fi%
  \global\let\svgwidth\undefined%
  \global\let\svgscale\undefined%
  \makeatother%
  \begin{picture}(1,0.51137885)%
    \lineheight{1}%
    \setlength\tabcolsep{0pt}%
    \put(0,0){\includegraphics[width=\unitlength,page=1]{figures/time_delay.pdf}}%
    \put(0.98321238,0.00756662){\color[rgb]{0,0,0}\makebox(0,0)[lt]{\lineheight{1.25}\smash{\begin{tabular}[t]{l}$t$\end{tabular}}}}%
    \put(0.27,0.15){$\underbrace{\hspace{34pt}}_{\textstyle \hat{t}_d}$}
    \put(0,0){\includegraphics[width=\unitlength,page=2]{figures/time_delay.pdf}}%
    \put(0.25243357,-0.01){\color[rgb]{0,0,0}\makebox(0,0)[lt]{\lineheight{1.25}\smash{\begin{tabular}[t]{l}${}^1t_i$\end{tabular}}}}%
    \put(0.38727193,-0.01){\color[rgb]{0,0,0}\makebox(0,0)[lt]{\lineheight{1.25}\smash{\begin{tabular}[t]{l}${}^2\bar{t}_i$\end{tabular}}}}%
    \put(0.67,0.42){\color[rgb]{0,0,0}\makebox(0,0)[lt]{\lineheight{1.25}\smash{\begin{tabular}[t]{l}${}^1{v(t)}$\end{tabular}}}}%
    \put(0.8,0.21){\color[rgb]{0,0,0}\makebox(0,0)[lt]{\lineheight{1.25}\smash{\begin{tabular}[t]{l}${}^2{v(t)}$\end{tabular}}}}%
  \end{picture}%
\endgroup%

%% file: text/gaussian.tex
\subsection{Gaussian Process Regression}
We take a GP regression approach to object trajectory estimation, leveraging the work in \cite{Anderson2015, Barfoot2017, Barfoot2014}, and apply it for object tracking.
The trajectory of an object is represented in continuous time and therefore we are able to query the solution at any time of interest.
By employing a special class of prior motion models with sparse structure, GP regression provides very efficient solutions \cite{Barfoot2017}.

We consider systems with a continuous-time GP model prior
\begin{equation}
\boldsymbol{x}(t) \sim \mathcal{GP}(\check{\boldsymbol{x}}(t), \check{\boldsymbol{P}}(t, t^{\prime})),
\end{equation}
and a discrete-time, linear measurement model:
\begin{equation}
\boldsymbol{y}_k(t) = \boldsymbol{C}_k \boldsymbol{x}_k(t_k) + \boldsymbol{n}_k,
\end{equation}
where $\boldsymbol{x}(t)$ is the state, $\check{\boldsymbol{x}}(t)$ is the mean function, $\check{\boldsymbol{P}}(t, t^{\prime})$ is the covariance function,
 $\boldsymbol{y}_k$ are the measurements, $\boldsymbol{n}_k \sim \mathcal{N}(\boldsymbol{0}, \boldsymbol{R}_k)$ is Gaussian measurement noise,
 and $\boldsymbol{C}_k$ is the measurement model matrix.
For now, we assume that the state is queried at the measurement times, and we will describe querying at other times later on.
Following the approach presented in \cite{Barfoot2017}, the Gaussian posterior evaluates to
\begin{multline}
p(\boldsymbol{x}|\boldsymbol{y}) = \mathcal{N} \Big ( \underbrace{ (\check{\boldsymbol{P}}^{-1} + \boldsymbol{C}^T \boldsymbol{R}^{-1} \boldsymbol{C})^{-1}
(\check{\boldsymbol{P}}^{-1}\check{\boldsymbol{x}} + \boldsymbol{C}^T \boldsymbol{R}^{-1}\boldsymbol{y} ) }_{ \hat{\boldsymbol{x}}, \text{ posterior mean}}, \\
\underbrace{ (\check{\boldsymbol{P}}^{-1} + \boldsymbol{C}^T \boldsymbol{R}^{-1} \boldsymbol{C})^{-1} \Big ) }_{\hat{\boldsymbol{P}}, \text{ posterior covariance}}.
\end{multline}
After rearranging the posterior mean expression, a linear system for $\hat{\boldsymbol{x}}$ is obtained
\begin{equation}
(\check{\boldsymbol{P}}^{-1} + \boldsymbol{C}^T \boldsymbol{R}^{-1} \boldsymbol{C}) \hat{\boldsymbol{x}}  =
(\check{\boldsymbol{P}}^{-1}\check{\boldsymbol{x}} + \boldsymbol{C}^T \boldsymbol{R}^{-1}\boldsymbol{y} ),
\label{eq:linsys}
\end{equation}
where $\check{\boldsymbol{P}}$, $\boldsymbol{C}$, and $\boldsymbol{R}$ are batch matrices defined as $\check{\boldsymbol{P}} = [ \check{\boldsymbol{P}}(t_i, t_j) ]_{ij}$, $\boldsymbol{C} = \text{diag}( \boldsymbol{C}_0, \dots, \boldsymbol{C}_N )$, and $\boldsymbol{R} = \text{diag} ( \boldsymbol{R}_0, \dots, \boldsymbol{R}_N )$, while $\hat{\boldsymbol{x}}$ and $\boldsymbol{y}$ are stacked vectors of prior states at measurement times and actual sensor measurements, $\hat{\boldsymbol{x}} = [ \hat{\boldsymbol{x}}(t_0), \dots, \hat{\boldsymbol{x}}(t_N) ]^T$ and $\boldsymbol{y} = [ \boldsymbol{y}_0, \dots, \boldsymbol{y}_N ]^T$, with $N$ being the number of measurements.
In general, time complexity for solving \eqref{eq:linsys}, as currently presented, is $\mathcal{O}(N^3)$ \cite{Anderson2015}.
Therefore, a special class of GP priors is introduced, whose sparsely structured matrices can be exploited to improve the computational efficiency.
\subsection{Exactly Sparse GP Priors}
The special class of GP priors is based on the following linear time-varying stochastic differential equation (LTV-SDE)
\begin{equation}
\dot{\boldsymbol{x}}(t) = \boldsymbol{F}(t) \boldsymbol{x}(t) + \boldsymbol{v}(t) + \boldsymbol{L}(t) \boldsymbol{w}(t),
\label{ltvsde}
\end{equation}
where $\boldsymbol{F}$ and $\boldsymbol{L}$ are system matrices, $\boldsymbol{v}$ is a known control input, and $\boldsymbol{w}(t)$ is generated by a white noise process.
The white noise process is itself a GP with zero mean value
\begin{equation}
\boldsymbol{w}(t) \sim \mathcal{GP} (\boldsymbol{0}, \boldsymbol{Q}_c \delta (t - t^{\prime})),
\end{equation}
where $\boldsymbol{Q}_c$ is a power spectral density matrix.

The mean and the covariance of the GP are generated from the solution of the LTV-SDE given in \eqref{ltvsde}
\begin{equation}
\check{\boldsymbol{x}} (t) = \boldsymbol{\Phi}(t, t_0) \check{\boldsymbol{x}}_0 + \int_{t_0}^t \boldsymbol{\Phi} (t, s) \boldsymbol{v}(s) \dd s,
\label{prior_x}
\end{equation}
\begin{multline}
\check{\boldsymbol{P}}(t, t^{\prime}) =  \boldsymbol{\Phi}(t, t_0) \check{\boldsymbol{P}}_{0} \boldsymbol{\Phi}(t^{\prime}, t_0)^T + \\ \int_{\text{t}_0}^{\text{min}(t,t^{\prime})} \boldsymbol{\Phi}(t, s)  \boldsymbol{L}(s) \boldsymbol{Q}_{c} \boldsymbol{L}(s)^T \boldsymbol{\Phi}(t^{\prime}, s)^T \dd s,
\label{prior_P}
\end{multline}
where $\check{\boldsymbol{x}}_0$ and $\check{\boldsymbol{P}}_0$ are the initial mean and covariance of the first state, and $\boldsymbol{\Phi} (t, s)$ is the state transition matrix \cite{Barfoot2014}.

Due to the Markov property of the LTV-SDE in \eqref{ltvsde}, the inverse kernel matrix $\check{\boldsymbol{P}}^{-1}$ of the prior, which is required for solving the linear system in \eqref{eq:linsys}, is exactly sparse block tridiagonal \cite{Barfoot2014}:
\begin{equation}
\check{\boldsymbol{P}}^{-1} = \boldsymbol{F}^{-T} \boldsymbol{Q}^{-1} \boldsymbol{F}^{-1},
\end{equation}
where
\begin{equation}
\boldsymbol{F}^{-1} =
\begin{bmatrix}
\boldsymbol{1} & 0 & ... & 0 & 0 \\
-\boldsymbol{\Phi}(t_1, t_0) & \boldsymbol{1} & ... & 0 & 0 \\
0 & -\boldsymbol{\Phi}(t_2, t_1) & \ddots & \vdots & \vdots \\
\vdots & \vdots & \ddots & \boldsymbol{1} & 0 \\
0 & 0 & ... & -\boldsymbol{\Phi}(t_N, t_{N-1}) & \boldsymbol{1}
\end{bmatrix}
\end{equation}
and
\begin{equation}
\boldsymbol{Q}^{-1} = \text{diag} (\check{\boldsymbol{P}}_0^{-1}, \boldsymbol{Q}_{0,1}^{-1}, ... , \boldsymbol{Q}_{N-1, N}^{-1})
\end{equation}
with
\begin{equation}
\boldsymbol{Q}_{a, b} = \int_{t_a}^{t_b} \boldsymbol{\Phi}(t_b, s) \boldsymbol{L}(s) \boldsymbol{Q}_c \boldsymbol{L}(s)^T \boldsymbol{\Phi}(t_b, s)^T \dd s.
\label{eq:Qab}
\end{equation}

This kernel allows for computationally efficient, structure-exploiting inference with $\mathcal{O}(N)$ complexity. This is the main advantage of the proposed exactly sparse GP priors based on a LTV-SDE in \eqref{ltvsde}.

As we previously stated, the key benefit of using GPs for the continuous-time object trajectory estimation is the possibility to query the state $\hat{\boldsymbol{x}}(\tau)$ at any time of interest $\tau$, and not only at measurement times.
For multisensor calibration, this proves to be extremely useful, since many sensors operate at different frequencies; thus, the GP approach enables us to temporally align the measurements.
If the prior proposed in \eqref{prior_x} is used, GP interpolation can be performed efficiently due to the aforementioned Markovian property of the LTV-SDE in (\ref{ltvsde}).
State $\hat{\boldsymbol{x}}(\tau)$ at $\tau \in [t_i, t_{i+1}]$ is a function of only its neighboring states \cite{Anderson2015},
\begin{equation}
\hat{\boldsymbol{x}}(\tau) = \check{\boldsymbol{x}}(\tau) + \boldsymbol{\Lambda}(\tau)(\hat{\boldsymbol{x}}_i - \check{\boldsymbol{x}}_i) + \boldsymbol{\Psi}(\tau)(\hat{\boldsymbol{x}}_{i+1} - \check{\boldsymbol{x}}_{i+1}),
\label{eq:intp1}
\end{equation}
\begin{equation}
\boldsymbol{\Lambda}(\tau) = \boldsymbol{\Phi} (\tau, t_i) - \boldsymbol{\Psi}(\tau)\boldsymbol{\Phi}(t_{i+1}, t_i),
\label{eq:intp2}
\end{equation}
\begin{equation}
\boldsymbol{\Psi}(\tau) = \boldsymbol{Q}_{i, \tau} \boldsymbol{\Phi}(t_{i+1}, \tau)^T \boldsymbol{Q}_{i, i+1}^{-1},
\label{eq:intp3}
\end{equation}
where $\boldsymbol{Q}_{a,b}$ is given in \eqref{eq:Qab}.
The fact that any state $\check{\boldsymbol{x}}(\tau)$ can be computed in $O(1)$ complexity can be exploited for efficient matching of trajectories of an object detected by multiple sensors.
The implementation details of the described method, as well as its utilization in the context of spatio-temporal calibration, are discussed in the sequel.

%% file: text/method.tex
The proposed calibration method can be separated in three consecutive steps: (i) representing the trajectories of moving objects captured by individual sensors with separate GPs, (ii) temporal calibration and correspondence registration based on GP interpolation, and (iii) extrinsic calibration using 3D point-to-point correspondence.
%
%
The method is specific in the sense that it relies on estimated moving object velocites.
The advantages of using velocity magnitudes are that (i) they are coordinate frame independed, in the sense that we can perform temporal calibration without prior extrinsic calibration, (ii) they can be used for efficient track association, and (iii) they are one-dimensional scalar values, thus making the optimization computationally less demanding.
Furthermore, using velocities has also an additional advantage with respect to position measurements.
Namely, different sensors can easily track different parts of moving objects, e.g., radars might track the vehicle's rear axle, while the LiDAR detects rear bumper, thus introducing bias in the position measurements.
By relying on the velocity magnitudes, we \emph{circmuvent the problem of position bias} from heterogenous sensor measurements.
A possible drawback is that temporal calibration requires changes in velocity to make the time delay observable; however, realistically, moving objects will accelerate or decelerate occasionally, thus offering enough information to perform calibration.
%

\subsection{GP Trajectory Representation}
\label{sec:method_gp}

For the calibration purposes, measurements from two individual sensors are used to create two separate GPs, where $s \in S = \{1,2\}$ represents the first and second sensor, respectively.
%
Since temporal calibration is unobservable during periods of constant velocity, we choose a GP motion prior of a higher order, namely constant acceleration, to capture the necessary maneuvering dynamics of the target
\begin{equation}
{}^s\boldsymbol{x}(t) =
\begin{bmatrix}
{}^s\boldsymbol{p}(t) \\
{}^s\boldsymbol{v}(t) \\
{}^s\boldsymbol{a}(t)
\end{bmatrix}
\sim
 \mathcal{GP}({}^s\check{\boldsymbol{x}}(t), {}^s\check{\boldsymbol{P}}(t, t^{\prime})).
\label{eq:traj_gaussian}
\end{equation}
To employ the mentioned constant acceleration motion prior, the LTV-SDE in \eqref{eq:linsys} has the following parameters
\begin{equation}
\boldsymbol{F}(t) =
\begin{bmatrix}
{0} && {1} && {0} \\
{0} && {0} && {1} \\
{0} && {0} && {0}
\end{bmatrix}
,
\boldsymbol{L}(t) =
\begin{bmatrix}
{0}  \\
{0}  \\
{1}
\end{bmatrix}
,
\boldsymbol{C}(t) =
\begin{bmatrix}
{1}  \\
{0}  \\
{0}
\end{bmatrix}^T
,
\end{equation}
while the matrices $\boldsymbol{\Phi}(t,s)$ and $\boldsymbol{Q}_{a, b}$ are defined as
\begin{equation}
\boldsymbol{\Phi}(t,s) =
\begin{bmatrix}
\boldsymbol{1} && (t-s)\boldsymbol{1} && \frac{(t-s)^2}{2}\boldsymbol{1} \\
\boldsymbol{0} && \boldsymbol{1} && (t-s)\boldsymbol{1} \\
\boldsymbol{0} && \boldsymbol{0} && \boldsymbol{1}
\end{bmatrix},
\end{equation}
\begin{equation}
\boldsymbol{Q}_{a, b}=
\begin{bmatrix}
\frac{{\Delta t}^5}{20}\boldsymbol{Q}_c && \frac{{\Delta t}^4}{8}\boldsymbol{Q}_c && \frac{{\Delta t}^3}{6}\boldsymbol{Q}_c \\
\frac{{\Delta t}^4}{8}\boldsymbol{Q}_c && \frac{{\Delta t}^3}{3}\boldsymbol{Q}_c && \frac{{\Delta t}^2}{2}\boldsymbol{Q}_c \\
\frac{{\Delta t}^3}{6}\boldsymbol{Q}_c && \frac{{\Delta t}^2}{2}\boldsymbol{Q}_c && {\Delta t}\boldsymbol{Q}_c
\end{bmatrix},
\end{equation}
with $\Delta t = t_b - t_a$.
\subsection{Temporal Calibration}
\label{sec:method_temporal}
%

In the proposed approach we isolate temporal from the extrinsic calibration.
Although some research advocates a unified approach to spatio-temporal calibration \cite{Furgale2013}, others claim that estimating uncorrelated quantities, such as time delay and homogeneous transforms, might degrade the final result \cite{Mair2011}.
Additional challenge in temporal calibration is computational complexity; namely, at each optimization step new correspondences need to be computed due to the new time delay perturbation.
Therefore, commonly we reduce the dimensionality of the problem and preferably remove correlation with the extrinsic calibration.
%
As previously discussed, our approach uses velocity magnitudes to reduce dimensionality and avoid the need for prior extrinsic calibration.

The temporal calibration process starts by choosing one of the sensors as the anchor, e.g., $s=1$.
Fig.~\ref{fig:temporal} illustrates velocity magnitude trajectories of the first and the second sensor (offsetted along the vertical axis for clarity), labeled ${}^1\boldsymbol{v}(t)$ and ${}^2\boldsymbol{v}(t)$, respectively.
%
%
The states of the anchor sensor at the measurement times, ${}^1t_i \in {}^1T, i=1 \ldots N_1$, are kept fixed for the correspondence registration.
At each step of iterative optimization, using the current estimate of the time delay $t_d$, states of the second sensor are interpolated using \eqref{eq:intp1}-\eqref{eq:intp3}.
The time delay is then estimated by minimizing the following criterion
\begin{equation}
\hat{t}_d=\underset{t_d}{\arg\min}\sum_{i=1}^{N_1} \bigg( ||{}^1{v}({}^1t_i)|| -||{}^2{v}({}^1t_i+t_d)|| \bigg)^2.
\label{eq:cost_temporal}
\end{equation}
It is worth noting that in the case of different sensor frame rates, the slower sensor should be chosen as the anchor to reduce the interpolation errors \cite{Huber2009}.
Finally, after the time delay optimization, we form a list of state correspondences between the states at measurement times ${}^1T$ of the anchor sensor, and corresponding interpolated times of the second sensor, ${}^2\tilde{T}$, defined as
\begin{equation}
{}^2\tilde{t}_i = {}^1t_i	+  \hat{t}_d , \quad {}^2\tilde{t}_i \in {}^2\tilde{T}, \quad i=1\ldots N_1.
\label{eq:time_aligned}
\end{equation}

The optimization problem is solved using iterative optimization; specifically, the Levenberg–Marquardt algorithm.
Since GP regression provides smooth trajectories and our motion prior includes acceleration estimation, we can derive the analytical Jacobian for each residual element in (\ref{eq:cost_temporal})
\begin{equation}
J_i=-\frac{{}^2\boldsymbol{v}({}^1t_i+t_d) \cdot {}^2\boldsymbol{a}({}^1t_i+t_d)}{||{}^2\boldsymbol{v}({}^1t_i+t_d)||},
\label{eq:jacobian}
\end{equation}
which simplifies the computational complexity of the optimization process.

\subsection{Extrinsic Calibration}
\label{sec:method_extrinsic}
%
%
The final estimate of the time delay is used to create a list of correspondences for the extrinsic calibration, where the continuous time representation enables elegant temporal registration.
%
The problem of aligning two sets of corresponding 3D points, i.e. 3D registration, is a widely studied problem and many closed-form solutions exist \cite{Eggert1997}, while the necessary condition for the identifiability is that we have at least three non-collinear points \cite{Horn1987}.
However, for the completeness of the approach we state that for extrinsic parameter estimation we solve the following problem
\begin{equation}
[\boldsymbol{\hat{\phi}},\boldsymbol{\hat{t}}] = \underset{\boldsymbol{\phi},\boldsymbol{t}}{\arg\min}\sum_{i=1}^{N_1} \bigg( {}^1\boldsymbol{p}({}^1t_i)
 - \boldsymbol{R}(\boldsymbol{\phi}) {}^2\boldsymbol{p}({}^2\tilde{t}_i) - \boldsymbol{t} \bigg)^2,
\label{eq:cost_position}
\end{equation}
where $\boldsymbol{\phi}$ and $\boldsymbol{t}$ are rotation and translation parameters between the sensors, respectively.

Note that ego-motion based calibration methods, which are not restricted to specific sensors and do not require specifically designed calibration targets, can be used for temporal and extrinsic calibration.
However, requiring that ego-motion is estimated by all the sensors in a complex system, solely for the purposes of sensor calibration, can be impractical.
For example, autonomous vehicles are commonly equipped with many cameras, LiDARs and radars where some sensors outperform the others in the ego-motion estimation.
On the other hand, in safety-critical navigation and dynamic scenarios detection and tracking of moving objects is usually performed by all the sensors, where both position and velocity have to be estimated.
Therefore, we believe that the proposed method has significant practical importance and paves the way to online tracking-based multisensor calibration.
Naturally, we do not dismiss a solution where calibration is performed by fusion of ego-motion and moving object tracking methods.
However, such an analysis is out of the scope of the current paper.

%% file: text/results.tex

\begin{figure}[t]
\centering
\input{figures/delay_histogram_comparison.tikz.tex}
	\vspace*{-0.5cm}
    \caption{Histograms compare estimated time delays using our method (\textit{GP}) to the method proposed in \cite{Huber2014}, where the results \textit{LIN-B} use all the measurements in a single dataset interval, while \textit{LIN-UB} uses only a subset which does not introduce a bias in the estimation. Ground truth time delay is $t_{d,gt}=0\,ms$.}
    \label{fig:delay_histogram}
\end{figure}
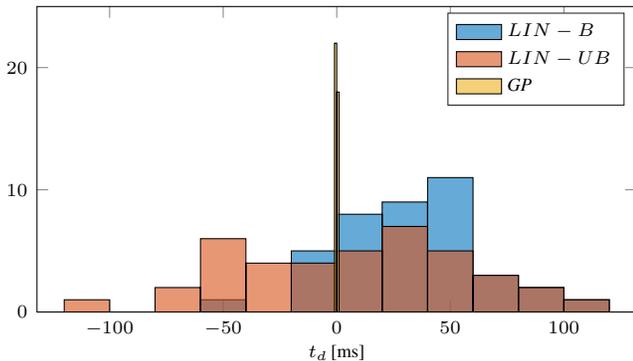

The proposed calibration method is tested in two real-world experiment and through simulations.
In the Sec.~\ref{sec:stereo} two synchronized global shutter image sensors are calibrated.
Section ~\ref{sec:mocap} provides details on calibration of a camera and a motion capture system to test heterogeneous sensors with separate clocks.
Sec.~\ref{sec:simulations} further examines the method through simulations, while the computational complexity of the method is reported in Sec.~\ref{sec:performance}.
In the experiments we have used a known target for convenience, even though the method does not rely on a special target.
We covered a planar board with an AprilTag \cite{Wang2016a}, a square fiducial marker of side length $a=14.35$\,cm, and motion capture markers.
%

%
%
%

\subsection{Stereo camera calibration}
\label{sec:stereo}

In this real-world experiment, we used two PointGrey BFLY-U3-23S6M-C global shutter image sensors combined with Kowa C-Mount $6$\,mm f/1.8-16 1" HC Series Fixed Lens of field of view $96.8{}^\circ \times 79.4{}^\circ$.
These cameras can be synchronized by an external trigger and the sampling rate of both cameras set to $t_m=0.05$\,s.
Even though such systems do not require temporal calibration, we have synchronized them to provide ground truth for the time delay.
We have recorded 40 1-minute datasets to compare the accuracy of the proposed temporal calibration method (labelled GP) to the two recent temporal calibration frameworks based on object tracking \cite{Huber2014,Fornaser2017}.
%
%

%
%
The first method, introduced in \cite{Fornaser2017}, correctly estimated 0-frame delay for all the 40 dataset we recorded, but it is limited to sample time resolution and requires the sensors operating at the same sampling rates.
The second method, proposed in \cite{Huber2014}, enables temporal calibration of asynchronous sensors in the continuous time domain.
However, this method relies on the position norm for the dimensionality reduction, which introduces significant errors with largely displaced sensors.
Additionally, we noticed that linear interpolation does not provide a smooth cost function, which is a prerequisite for gradient based iterative optimization methods (e.g. Levenberg-Marquardt method).
%
%
To illustrate how the dataset can cause significant bias in the estimation, we tested the second method on full intervals and their subsets.
Namely, in the first third of our 1-minute intervals, the target is moved along the optical axis of both cameras and the method provides unbiased estimates (labelled \textit{LIN-UB}) with distribution $\hat{t}_d \sim \mathcal{N}(\SI{0.006} \,\mathrm{s},\,\SI{2.6e-3})$.
On the other hand, for the remaining parts of the intervals, motion is not aligned with the optical axis and the position norm measurements differ for the two cameras as they are displaced far enough.
Therefore, we believe that the position norm is not the most appropriate dimensionality reduction technique as it is not frame-invariant.
The results for the full interval (labelled \textit{LIN-B}) follow the distribution $\hat{t}_d \sim \mathcal{N}(\SI{0.0322} \,\mathrm{s},\,\SI{1.1e-3})$. 
Additionally, a similar variance on the time delay is reported in \cite{Huber2014} for the calibration of a camera and a motion capture system.
%

%
%
The proposed method is able to produce an unbiased time delay estimate.
%
The estimated Gaussian distribution from the results is $\hat{t}_d \sim \mathcal{N}(\SI{-6.85e-5} \,\mathrm{s},\,\SI{1.88e-7})$.
All estimates were within the range $(-0.82,0.83)$\,ms which corresponds to the $\pm 1.7\%$ range of the sampling interval $t_m$.
We can see that the proposed method outperforms other algorithms in both accuracy and precision.
Fig 2 illustrates the advantage of the proposed method over the position norm method \cite{Huber2014}.

\begin{figure*}[t]
\centering
\subfloat[][Wide preview illustrating local minima and global minimum.]{
\hspace*{-0.5cm}
\input{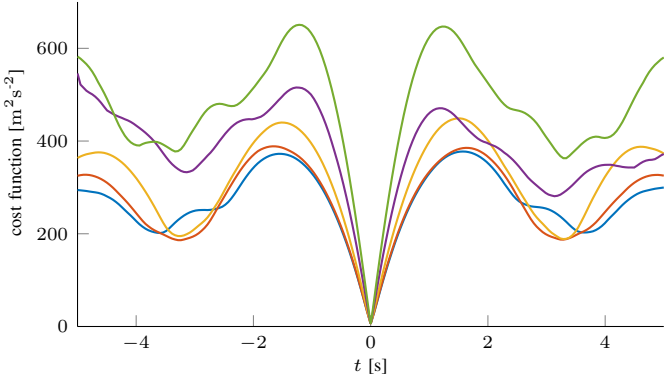}
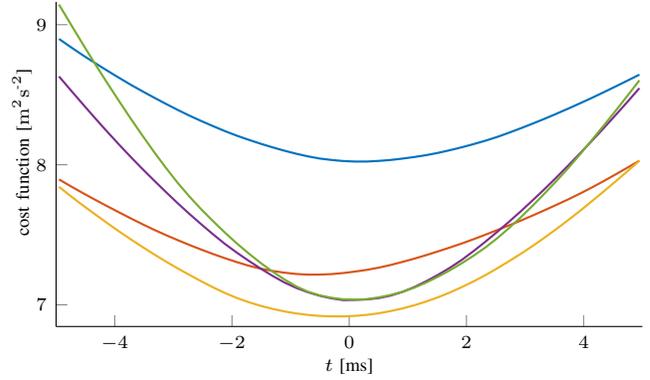
\label{fig:cost_big}}
\subfloat[][Closer view around the ground truth.]{
\hspace*{0.3cm}
\input{figures/delay_cost_small_5_reduced_ms.tikz.tex}
\label{fig:cost_small}}
\caption{Cost function of the proposed calibration method (GP) for 5 different experiments.}
\label{fig:costs}
\end{figure*}

To further gain insight in the proposed temporal calibration method, we examined the cost function defined by \eqref{eq:cost_temporal}.
Fig.~\ref{fig:cost_big} shows the value of the cost function in the interval $t_d \in (-5,-5)$\,s, while Fig.~\ref{fig:cost_small} provides a closer look around the global optimum, $t_d \in (-5,-5)$\,ms.
For clarity, only 5 out of 40 experiments are shown, while the rest follow the same pattern.
From Fig.~\ref{fig:cost_big}, we can see that the cost function has many local minima, while the global minimum always resides near the ground truth.
Since our method uses an iterative solver, proper initialization is necessary.
We can see that initializing the time delay to a starting point in the interval $t_{d,0} \in (-1,1)$\,s would enable the method to converge to the global minimum for all the experiments.
The local minima are tightly coupled with the executed target motion and can be further spread from the global minimum by avoiding repetitive motion or increasing its period.
%
%
Figure~\ref{fig:cost_small} shows that our cost function is smooth with a minimum around the ground truth value, thus enabling stable and accurate results using an iterative optimization.

After using the estimated time delay to perform interpolation of the corresponding measurements, we apply 3D point-to-point registration for extrinsic calibration. 
Mean of the estimated Euler angles was [$-0.37$,$-0.25$,$-0.21$]$\,{}^{\circ}$ with an approximate accuracy of $0.3{}^{\circ}$, while the estimated mean translation was [$569.5$,$3.1$,$-9.7$]\,mm with an approximate accuracy of 8\,mm.

\subsection{Camera--Motion Capture calibration}
\label{sec:mocap}
We calibrated a single camera from the previous experiment with the Optitrack motion capture system.
The Optitrack motion capture provides 3D position measurements at a significantly higher frame rate of 120\,Hz, while it creates timestamps using a dedicated computer's local clock and transmits the data over the wireless network.
Given that, this setup gives us two options regarding data timestamps: (i) to use the measurement arrival time at the central computer and (ii) to use local timestamps provided by each sensor.
Both options were analysed in a 2-hour experiment with 1-minute intervals recorded every 30 minutes and we discuss the results in the sequel.

%

In the first approach, arrival times at the central computer were used, and the proposed method yielded estimated time delays within a range $t_d = [51.8,57.9]$\,ms.
The estimated time delays are consistent up to several miliseconds (6.1\,ms), while the mean value can be interpreted as the average network introduced delay (mean value was 54.5\,ms).
These results are particularly useful for online sensor fusion as they provide the data communication lag assessment.
However, it should be noted that the delay estimated using arrival times can differ significantly during the experiments due to changing intensity of network traffic or other protocol induced stochastic effects.

In the second approach, we used local timestamps provided by individual sensor clocks.
Even though this approach eliminates stochastic effects associated with communication over a network, different rates of  individual sensor clocks can cause drift in the time delay estimation.
This effect is present in the estimated time delays which are listed sequentially from the start to the end of the experiment: $t_d = [-613.7,-712.3,-825.6,-911.3,-998.4]$\,ms.
Given that, we can notice a time delay drift of the motion capture local computer with respect to the central computer of $-53.7$\,$\mu$s/s.
To further corroborate this result, we compare the local motion capture timestamps to the arrival times at the central computer and we obtained a clock drift estimate of $-53.2$\,$\mu$s/s which confirms our result.
%
Finally, we report that the estimated extrinsic parameters have a maximum deviation of 0.065 deg in Euler angles and 3 mm in the translation with respect to the mean estimated value.

\subsection{Simulation experiment}
\label{sec:simulations}

Synthetic dataset in the simulation consisted of $N=500$ 1-minute long trajectories.
Trajectories were defined in the global frame, while we sampled the measurements with an interval of $t_m=0.05$\,s in  each sensors local frame.
To test the GP interpolation, two sensors sampled the trajectory in a counter-phase, i.e., the second sensor sampled the trajectory with half the sampling interval after the first.
Additionally, we have put an offset on the starting point of trajectory sampling to $t_s=0.1$\,s, i.e., the second sensor started sampling 2.5 sampling intervals after the first.
Local timestamps of individual sensors started from $t_0=0$\,s, resulting in a ground truth delay $\bar{t}_d =  t_s + t_m/2 = 0.125$\,s.
White noise with covariance matrix $R=\mathrm{diag}(\sigma^2,\sigma^2,\sigma^2), \sigma=0.01$\,m was added to the measurements resembling the noise found in the real-world experiment.
Ground truth extrinsic calibration was given in Euler angles $\boldsymbol{\bar{\Theta}} = [ \bar{\Theta}_z, \bar{\Theta}_y, \bar{\Theta}_x] = [45{}^\circ, 20{}^\circ, 0{}^\circ]$ and translation $\boldsymbol{\bar{t}} = [ \bar{t}_x, \bar{t}_y, \bar{t}_z] = [1.0$\,m$, -1.0$\,m$,1.0$\,m$]$.

Temporal calibration converged to the ground truth value with a Gaussian distribution of the estimation error $(\hat{t}_d-\bar{t}_d) \sim \mathcal{N}(\SI{-1.06e-5}\,\mathrm{s}\,,\SI{7.67e-7})$.
%
Estimated values are within the range of $(-1.5,1.5)$\,ms interval which implies the precision of $ \pm 3\% $ of the sampling interval.
Finally, the estimated extrinsic parameters have a maximum deviation of 0.1 deg in Euler angles and 3 mm in the translation with respect to the mean estimated value.

\subsection{Execution performance}
\label{sec:performance}

Execution performance of the proposed method was tested on 40 datasets from Sec.~\ref{sec:stereo} on a PC with a i7-6700HQ CPU at 2.6\,GHz $\times$ 8 and 16\,GB of 2133\,MHz DDR4 RAM.
The calibration starts with two separate GP regressions for individual sensors that are completely decoupled and performed in separate threads.
On average, 1-minute intervals consisted of 1138 measurements requiring $t_{GP}=49$\,ms for the complete GP regression.
In our implementation, we handle missing measurements and varying sample times.
Under the assumption of constant sample times and absence of missing measurements, further improvements of GP regression performance are possible through offline construction of required batch matrices.
After the GP regression, we used the Ceres solver \cite{ceres-solver} to compute the time delay estimate.
On average, 1-minute trajectory pairs were represented by GP's generated 1087 correspondences.
When the time delay was initialized as $t_d = 0.5$\,s, it took around 6 iterations to converge, which translates to the average optimization time of $t_{opt}=45$\,ms.
Finally, the total time required for the delay estimation was on average $t_{total} = t_{GP} + t_{opt} = 94$\,ms.
It is important to mentions that the algorithm time complexity is $O(n)$ which makes the method scalable for fast sensors and longer time intervals.

%% file: figures/delay_histogram_comparison.tikz.tex
%
%
\definecolor{mycolor1}{rgb}{0.00000,0.44700,0.74100}%
\definecolor{mycolor2}{rgb}{0.85000,0.32500,0.09800}%
\definecolor{mycolor3}{rgb}{0.92900,0.69400,0.12500}%
\begin{tikzpicture}

\begin{axis}[%
width=0.9\columnwidth,
height=1.6in,
at={(2.512in,1.284in)},
scale only axis,
xmin=-132,
xmax=132,
ymin=0,
ymax=25,
x label style={at={(axis description cs:0.5,0.05)},anchor=north},
xlabel={$t_d$\,[ms]},
axis background/.style={fill=white},
legend style={legend cell align=left, align=left, draw=white!15!black}
]
\addplot[ybar interval, fill=mycolor1, fill opacity=0.6, draw=black, area legend] table[row sep=crcr] {%
x	y\\
-60	1\\
-40	0\\
-20	5\\
0	8\\
20	9\\
40	11\\
60	3\\
80	2\\
100	1\\
120	1\\
};
\addlegendentry{\textit{$LIN-B$}}

\addplot[ybar interval, fill=mycolor2, fill opacity=0.6, draw=black, area legend] table[row sep=crcr] {%
x	y\\
-120	1\\
-100	0\\
-80	2\\
-60	6\\
-40	4\\
-20	4\\
0	5\\
20	7\\
40	5\\
60	3\\
80	2\\
100	1\\
120	1\\
};
\addlegendentry{\textit{$LIN-UB$}}

\addplot[ybar interval, fill=mycolor3, fill opacity=0.6, draw=black, area legend] table[row sep=crcr] {%
x	y\\
-1	22\\
0	18\\
1	18\\
};
\addlegendentry{\textit{GP}}

\end{axis}
\end{tikzpicture}%

%% file: figures/delay_cost_small_5_reduced_ms.tikz.tex
%
%
\definecolor{mycolor1}{rgb}{0.00000,0.44700,0.74100}%
\definecolor{mycolor2}{rgb}{0.85000,0.32500,0.09800}%
\definecolor{mycolor3}{rgb}{0.92900,0.69400,0.12500}%
\definecolor{mycolor4}{rgb}{0.49400,0.18400,0.55600}%
\definecolor{mycolor5}{rgb}{0.46600,0.67400,0.18800}%
\begin{tikzpicture}

\begin{axis}[%
width=0.88\columnwidth,
height=1.7in,
at={(2.512in,1.284in)},
scale only axis,
xmin=-5,
xmax=5,
ymin=6.84520521936062,
ymax=9.1626338082278,
x label style={at={(axis description cs:0.5,0.05)},anchor=north},
y label style={at={(axis description cs:0.1,0.25)},anchor=west},
ylabel={cost function [$\mathrm{m}^2\mathrm{s}^{\text{-}2}$]},
xlabel={$t$ [ms]},
axis background/.style={fill=white},
axis x line*=bottom,
axis y line*=left
]
\addplot [color=mycolor1, forget plot, thick]
  table[row sep=crcr]{%
-4.94949494949495	8.89847668172824\\
-4.74747474747475	8.83967648344179\\
-4.54545454545454	8.7826576820146\\
-4.34343434343434	8.72853057468981\\
-4.14141414141414	8.67569886272512\\
-3.93939393939394	8.62429491838123\\
-3.73737373737374	8.5740137423308\\
-3.53535353535354	8.52652615862157\\
-3.33333333333333	8.48071716428376\\
-3.13131313131313	8.43597858612116\\
-2.92929292929293	8.39255642990875\\
-2.72727272727273	8.35131703798526\\
-2.52525252525253	8.31259638766571\\
-2.32323232323232	8.27605055678326\\
-2.12121212121212	8.24068517999439\\
-1.91919191919192	8.20804156572445\\
-1.71717171717172	8.17869969939661\\
-1.51515151515152	8.15087394980314\\
-1.31313131313131	8.12514264495119\\
-1.11111111111111	8.10124237225985\\
-0.90909090909091	8.07928813360351\\
-0.707070707070706	8.06026915671889\\
-0.505050505050505	8.04461844347699\\
-0.303030303030303	8.03438348652359\\
-0.101010101010102	8.02657821842883\\
0.101010101010102	8.022990202557\\
0.303030303030303	8.02311100217577\\
0.505050505050505	8.02757573277423\\
0.707070707070708	8.0343526117408\\
0.90909090909091	8.04411658019007\\
1.11111111111111	8.05548808403242\\
1.31313131313131	8.06870625880961\\
1.51515151515152	8.08504345737904\\
1.71717171717172	8.10410253014742\\
1.91919191919192	8.12477210286612\\
2.12121212121212	8.14752013115871\\
2.32323232323232	8.17186829490106\\
2.52525252525252	8.19885646695567\\
2.72727272727273	8.2291042146889\\
2.92929292929293	8.26110663650799\\
3.13131313131313	8.29434757417327\\
3.33333333333333	8.32855218662137\\
3.53535353535354	8.36447187742263\\
3.73737373737374	8.40138523261242\\
3.93939393939394	8.43929519948319\\
4.14141414141414	8.47809599679257\\
4.34343434343434	8.51836081839395\\
4.54545454545454	8.55950283884903\\
4.74747474747475	8.6010947340887\\
4.94949494949495	8.64346355698363\\
};
\addplot [color=mycolor2, forget plot, thick]
  table[row sep=crcr]{%
-4.94949494949495	7.89552940368168\\
-4.74747474747475	7.8460691686081\\
-4.54545454545454	7.79843030059208\\
-4.34343434343434	7.75238339330075\\
-4.14141414141414	7.7070927533703\\
-3.93939393939394	7.66313447768893\\
-3.73737373737374	7.61973991258723\\
-3.53535353535354	7.57743476030307\\
-3.33333333333333	7.53670223489642\\
-3.13131313131313	7.49868502103188\\
-2.92929292929293	7.46256223985141\\
-2.72727272727273	7.42766827101522\\
-2.52525252525253	7.39452303399117\\
-2.32323232323232	7.3627579157162\\
-2.12121212121212	7.33309328907158\\
-1.91919191919192	7.30513967701481\\
-1.71717171717172	7.27942587077372\\
-1.51515151515152	7.25844968143388\\
-1.31313131313131	7.24222601092078\\
-1.11111111111111	7.22983600307767\\
-0.90909090909091	7.22036335992191\\
-0.707070707070706	7.21551732177158\\
-0.505050505050505	7.21570228102562\\
-0.303030303030303	7.21936487764921\\
-0.101010101010102	7.22692424317093\\
0.101010101010102	7.23744803994571\\
0.303030303030303	7.24988954919541\\
0.505050505050505	7.26531923063316\\
0.707070707070708	7.28450467442592\\
0.90909090909091	7.30650138939199\\
1.11111111111111	7.32962920285331\\
1.31313131313131	7.35489969029391\\
1.51515151515152	7.38168630888112\\
1.71717171717172	7.40919777910301\\
1.91919191919192	7.43775422921085\\
2.12121212121212	7.46847816950557\\
2.32323232323232	7.50066164589964\\
2.52525252525252	7.53376508506792\\
2.72727272727273	7.56739868110988\\
2.92929292929293	7.60177386495139\\
3.13131313131313	7.63810302303807\\
3.33333333333333	7.6756222528729\\
3.53535353535354	7.71400862377668\\
3.73737373737374	7.75316868721187\\
3.93939393939394	7.79475754240614\\
4.14141414141414	7.83826464575397\\
4.34343434343434	7.88412495461743\\
4.54545454545454	7.93169939007868\\
4.74747474747475	7.9799594519907\\
4.94949494949495	8.02917458098492\\
};
\addplot [color=mycolor3, forget plot, thick]
  table[row sep=crcr]{%
-4.94949494949495	7.84237690221478\\
-4.74747474747475	7.7751563216977\\
-4.54545454545454	7.71078316097481\\
-4.34343434343434	7.64890107123978\\
-4.14141414141414	7.58772195150467\\
-3.93939393939394	7.52797875299109\\
-3.73737373737374	7.47071338367164\\
-3.53535353535354	7.41555838865206\\
-3.33333333333333	7.36189664629818\\
-3.13131313131313	7.31011796291035\\
-2.92929292929293	7.26095728865376\\
-2.72727272727273	7.21394134631728\\
-2.52525252525253	7.1683125344078\\
-2.32323232323232	7.12502532860717\\
-2.12121212121212	7.08469789126016\\
-1.91919191919192	7.0490918233319\\
-1.71717171717172	7.01916614147055\\
-1.51515151515152	6.99305216732233\\
-1.31313131313131	6.9715484758971\\
-1.11111111111111	6.9532607600915\\
-0.90909090909091	6.93877082731615\\
-0.707070707070706	6.92799788287639\\
-0.505050505050505	6.92067717205772\\
-0.303030303030303	6.91742243495106\\
-0.101010101010102	6.91733663778254\\
0.101010101010102	6.9212169169007\\
0.303030303030303	6.92835454312121\\
0.505050505050505	6.93957968403689\\
0.707070707070708	6.95539954547308\\
0.90909090909091	6.97415486242161\\
1.11111111111111	6.99711894868006\\
1.31313131313131	7.02342921362906\\
1.51515151515152	7.05384192134579\\
1.71717171717172	7.08809968888578\\
1.91919191919192	7.12626849276576\\
2.12121212121212	7.16734965288574\\
2.32323232323232	7.21199724165419\\
2.52525252525252	7.26038785049805\\
2.72727272727273	7.31182030982237\\
2.92929292929293	7.36483339687804\\
3.13131313131313	7.42083918547357\\
3.33333333333333	7.47999149327234\\
3.53535353535354	7.54162075223881\\
3.73737373737374	7.60560243035883\\
3.93939393939394	7.67165346227183\\
4.14141414141414	7.7402438086564\\
4.34343434343434	7.81061496052978\\
4.54545454545454	7.88181890676012\\
4.74747474747475	7.95430086557451\\
4.94949494949495	8.02857664239911\\
};
\addplot [color=mycolor4, forget plot, thick]
  table[row sep=crcr]{%
-4.94949494949495	8.63056765482172\\
-4.74747474747475	8.53133372363147\\
-4.54545454545454	8.4335736256084\\
-4.34343434343434	8.33698684241429\\
-4.14141414141414	8.24249968353639\\
-3.93939393939394	8.1503190399492\\
-3.73737373737374	8.06047669413599\\
-3.53535353535354	7.97376779500881\\
-3.33333333333333	7.88973536431212\\
-3.13131313131313	7.80721253134739\\
-2.92929292929293	7.72682702857112\\
-2.72727272727273	7.64960443579908\\
-2.52525252525253	7.57557623763208\\
-2.32323232323232	7.50381149444826\\
-2.12121212121212	7.43513217080997\\
-1.91919191919192	7.37316769614177\\
-1.71717171717172	7.31506467974624\\
-1.51515151515152	7.25990079084149\\
-1.31313131313131	7.20801046066637\\
-1.11111111111111	7.16073802918573\\
-0.90909090909091	7.12170944585746\\
-0.707070707070706	7.09047422135209\\
-0.505050505050505	7.06478653120635\\
-0.303030303030303	7.04619034234213\\
-0.101010101010102	7.03326689947692\\
0.101010101010102	7.03484086979789\\
0.303030303030303	7.03849527424541\\
0.505050505050505	7.04783804802462\\
0.707070707070708	7.06570386785923\\
0.90909090909091	7.0922298147671\\
1.11111111111111	7.12707224378789\\
1.31313131313131	7.16827951834505\\
1.51515151515152	7.21401506422711\\
1.71717171717172	7.26441949146106\\
1.91919191919192	7.32039580471447\\
2.12121212121212	7.38363286649978\\
2.32323232323232	7.45080094382493\\
2.52525252525252	7.52090911681732\\
2.72727272727273	7.59349746274361\\
2.92929292929293	7.66876597758285\\
3.13131313131313	7.74717999879078\\
3.33333333333333	7.826989997962\\
3.53535353535354	7.91018854823523\\
3.73737373737374	7.9954036142566\\
3.93939393939394	8.08254622234348\\
4.14141414141414	8.17184300491382\\
4.34343434343434	8.26262518684471\\
4.54545454545454	8.35554830138798\\
4.74747474747475	8.45001109696166\\
4.94949494949495	8.54689171154694\\
};
\addplot [color=mycolor5, forget plot, thick]
  table[row sep=crcr]{%
-4.94949494949495	9.14343322502544\\
-4.74747474747475	9.00197137041351\\
-4.54545454545454	8.86278309997871\\
-4.34343434343434	8.72562901434622\\
-4.14141414141414	8.59305787545877\\
-3.93939393939394	8.46341473050643\\
-3.73737373737374	8.33626596686624\\
-3.53535353535354	8.21346784546997\\
-3.33333333333333	8.09415449045183\\
-3.13131313131313	7.9806467902394\\
-2.92929292929293	7.87260669903895\\
-2.72727272727273	7.77402619313249\\
-2.52525252525253	7.68257898060429\\
-2.32323232323232	7.59763247191532\\
-2.12121212121212	7.51607581920252\\
-1.91919191919192	7.43898414492319\\
-1.71717171717172	7.3677466785421\\
-1.51515151515152	7.30025752516777\\
-1.31313131313131	7.23888308273818\\
-1.11111111111111	7.18262409413557\\
-0.90909090909091	7.13268582482302\\
-0.707070707070706	7.09410056635895\\
-0.505050505050505	7.06777222274251\\
-0.303030303030303	7.05031965251647\\
-0.101010101010102	7.03955280805669\\
0.101010101010102	7.03697540245375\\
0.303030303030303	7.04070282629376\\
0.505050505050505	7.05318496649212\\
0.707070707070708	7.07053249778169\\
0.90909090909091	7.09310389486664\\
1.11111111111111	7.12231380143381\\
1.31313131313131	7.15917184168414\\
1.51515151515152	7.20087833054793\\
1.71717171717172	7.24647990233919\\
1.91919191919192	7.29643489952049\\
2.12121212121212	7.35065972404247\\
2.32323232323232	7.40945490986896\\
2.52525252525252	7.47543503595335\\
2.72727272727273	7.54828369538094\\
2.92929292929293	7.62613225930827\\
3.13131313131313	7.70854637421841\\
3.33333333333333	7.79532709691486\\
3.53535353535354	7.88445882605329\\
3.73737373737374	7.97700744317759\\
3.93939393939394	8.07563118161707\\
4.14141414141414	8.17769605614934\\
4.34343434343434	8.28108437123825\\
4.54545454545454	8.38620945254711\\
4.74747474747475	8.49378195336886\\
4.94949494949495	8.60279725697943\\
};
\end{axis}
\end{tikzpicture}%

%% file: text/conclusion.tex
In this paper we have proposed a spatio-temporal multisensor calibration method based on Gaussian processes estimated moving object tracking.
The proposed method relies on the estimated object velocities, where velocity magnitudes are used for time delay estimation, since they are coordinate frame independent and can be used before the sensors have been extrinsically calibrated.
Thereafter, using the obtained time delay the extrinsic parameters are estimated based on 3D point correspondences.
The method is applicable to any multisensor setup, as long as sensors can estimate the 3D position of a moving object.
We have validated the proposed calibration method in simulations and real-world experiments on two multisensor setups. The first setup consisted of two externally triggered cameras, thus having ground truth readily available, while the second setup consisted of a camera and a motion capture system.
We also compared the proposed approach to two state-of-the-art methods and the results showed that the proposed method outperfomed other approaches and that it reliably estimates the time delay up to a fraction of the sampling rate of the fastest sensor.
The proposed method bears the potential to serve as the base for on-line calibration of autonomous vehicle or robot heterogeneous sensors by tracking moving objects in the environment -- an information that is already available in most autonomous systems navigating in dynamic environments.

\balance